\documentclass[letterpaper]{article} 
\usepackage{aaai24}  
\usepackage{times}  
\usepackage{helvet}  
\usepackage{courier}  
\usepackage[hyphens]{url}  
\usepackage{graphicx} 
\usepackage{natbib}  
\usepackage{caption} 
\usepackage{algorithm}
\usepackage{algorithmic}
\usepackage{xcolor}

\usepackage{newfloat}
\usepackage{listings}
\DeclareCaptionStyle{ruled}{labelfont=normalfont,labelsep=colon,strut=off} 
\floatstyle{ruled}
\newfloat{listing}{tb}{lst}{}
\floatname{listing}{Listing}
\usepackage{framed}
\usepackage{multirow}
\usepackage{pifont}
\usepackage{bm}
\usepackage{threeparttable}
\usepackage{times}
\usepackage{latexsym}
\usepackage{amsmath}
\usepackage{graphicx}
\usepackage{array}
\usepackage{multirow} 
\usepackage{tabularx} 

\title{LLMEval: A Preliminary Study on How to Evaluate Large Language Models}
\author{
    Yue Zhang\textsuperscript{\rm 1}\equalcontrib,
    Ming Zhang\textsuperscript{\rm 1}\equalcontrib,
    Haipeng Yuan\textsuperscript{\rm 1},
    Shichun Liu\textsuperscript{\rm 1},
    Yongyao Shi\textsuperscript{\rm 3}\\
    Tao Gui\textsuperscript{\rm 2},
    Qi Zhang\textsuperscript{\rm 1}\thanks{Corresponding author.},
    Xuanjing Huang\textsuperscript{\rm 1}
}

\affiliations{
    \textsuperscript{\rm 1} School of Computer Science, Fudan University, Shanghai, China\\
    \textsuperscript{\rm 2} Institute of Modern Languages and Linguistics, Fudan University, Shanghai, China\\
    \textsuperscript{\rm 3} Shanghai Advanced Institute of Finance, Shanghai Jiaotong University, Shanghai, China\\

    yuezhang.fdu@gmail.com, mingzhang23@m.fudan.edu.cn, 
    \{fdyhp49,liusc2020\}@gmail.com, yyshi.23@saif.sjtu.edu.cn\\
    \{tgui,qz,xjhuang\}@fudan.edu.cn
}
\usepackage{bibentry}

\begin{document}

\maketitle

\begin{abstract}

Recently, the evaluation of Large Language Models has emerged as a popular area of research. 
The three crucial questions for LLM evaluation are ``what, where, and how to evaluate''.
However, the existing research mainly focuses on the first two questions, which are basically what tasks to give the LLM during testing and what kind of knowledge it should deal with.
As for the third question, which is about what standards to use, the types of evaluators, how to score, and how to rank, there hasn't been much discussion.
In this paper, we analyze evaluation methods by comparing various criteria with both manual and automatic evaluation, utilizing onsite, crowd-sourcing, public annotators and GPT-4, with different scoring methods and ranking systems. 
We propose a new dataset, LLMEval and conduct evaluations on 20 LLMs. 
A total of 2,186 individuals participated, leading to the generation of 243,337 manual annotations and 57,511 automatic evaluation results.
We perform comparisons and analyses of different settings and conduct 10 conclusions that can provide some insights for evaluating LLM in the future. The dataset and the results are publicly available at 
\url{https://github.com/llmeval} .

\end{abstract}

\section{Introduction}
In recent years, Large Language Models (LLMs) have emerged as a highly significant and extensively explored area of research.
As the capabilities of these LLMs continue to advance, it becomes increasingly crucial to assess their performance and understand their limitations.
However, traditional metrics for generative models, for example, BLEU\cite{BLEU}, ROUGE\cite{ROUGE}, WMD\cite{WMD}, MoverScore\cite{MoverScore}, can only capture one or a few aspects of the model's capabilities.

\begin{table}
\centering
\begin{threeparttable}
\fontsize{9}{10}\selectfont
\begin{tabular}{
lccccc
}
\hline
    \rule{0pt}{3.1cm}\textbf{Studies} &  
    \rotatebox{90}{\textbf{Automatic}} & 
    \rotatebox{90}{\textbf{Onsite Annotator\tnote{\dag}}}  &  
    \rotatebox{90}{\textbf{Crowd-sourcing\tnote{\dag}}}  & 
    \rotatebox{90}{\textbf{Public Annotator\tnote{\dag}}}  \\

\hline
HELM\cite{HELM}                         &   \ding{51} &      {}     &      {}     &      {}     \\
MMLU\cite{MMLU}                         &   \ding{51} &      {}     &      {}     &      {}     \\
C-Eval\cite{huang2023ceval}             &   \ding{51} &      {}     &      {}     &      {}     \\
AGIEval\cite{AGIEval}                   &   \ding{51} &      {}     &      {}     &      {}     \\
BERTScore\cite{BERTScore}                   &   \ding{51} &      {}     &      {}     &      {}     \\
AlpacaFarm\cite{AlpacaFarm}             &   \ding{51} &      {}     &   \ding{51} &      {}     \\
Chatbot Arena\cite{zheng2023judging}    &   \ding{51} &      {}     &      {}     &   \ding{51} \\
\textbf{Ours\tnote{\ddag}}              &   \ding{51} &  \ding{51}  &   \ding{51} &   \ding{51} \\
\hline
\end{tabular}
 \begin{tablenotes}
      \item[\textbf{\dag}] Manual evaluation with different types of annotators
      \item[\textbf{\ddag}] Despite the \textit{type of annotator}, our study also addresses the problems of \textit{what criteria to use, how to score and how to rank}.
    \end{tablenotes}
\end{threeparttable}
\caption{Evaluation Methods employed in LLM Evaluations}
\label{tab:studies}
\end{table}

Recent research has started to explore the measurement of LLM from a more synthesized perspective.
Those studies can be divided into two categories, automatic and manual evaluation, based on whether scores can be automatically calculated.
There have been numerous efforts to carry out automatic evaluation.
HELM\cite{HELM} achieves synthesized evaluation by combining a large number of existing datasets.
MMLU\cite{MMLU} employs multiple-choice questions for automated evaluation. 
C-Eval\cite{huang2023ceval} is a Chinese benchmark similar to MMLU.
AGIEval\cite{AGIEval} utilizes both cloze tasks and multi-choice question-answering tasks simultaneously.
Approaches like BERTScore\cite{BERTScore} assign scores to outputs of LLMs by employing another LLM.
As the capabilities of LLMs increasingly strengthen, apart from automated evaluations, manual evaluations are also an option.
ChatBot Arena\cite{zheng2023judging} allows public evaluator vote between two LLMs to rate them.
AlpacaFarm\cite{AlpacaFarm} leverages API LLMs to mimic manual evaluations as a low-cost replacement.

In a recent survey \cite{Survey_What_Where_How}, three questions are raised about LLM evaluation, ``what, where and how to evaluate''.
``What to evaluate'' is about determining the tasks for the LLMs to execute during evaluation.
``Where to evaluate'' discusses the knowledge domains in which to evaluate the LLMs.
These two questions have been quite extensively discussed.
However, there's less research on "how to evaluate," which refers to the specific methods for evaluation. This includes scoring criteria, grading approaches, ranking systems, and the type of annotators to use if manual evaluation is employed.

In this paper, we focus on ``how to evaluate''.
As shown in Table \ref{tab:studies}, our study examines both manual evaluation and GPT-4 based automatic evaluation.
Compared to LLM-as-a-Judge \cite{zheng2023judging}, our study employs a greater number of annotator types in manual evaluation.
Besides that, we also compare various scoring criteria, grading methods, and ranking systems.
In total, we gathered 243,337 manual annotations and 57,511 automatic evaluation results.
We will release all the annotated data to Github when the anonymity period ends.

In general, when considering how to conduct an evaluation of an LLM, we come across three crucial questions that need to be addressed.
 
\textbf{Q1: Which criteria should we take into account when evaluating LLMs?}
We can judge an LLM from various angles, like how accurate and fluent its answers are. 
But are all these criteria really needed? 
Could there be some aspects where all current LLMs have already done well enough, so further evaluation might not be necessary?

We conduct a comparison to the five criteria, accuracy, informativeness, fluency, logical coherence and harmlessness.
The results show that across various criteria, existing LLMs all have demonstrated notable performance in terms of harmlessness. The differentiating factors lie in the metrics of informativeness and accuracy.

\textbf{Q2: Which annotation methods should be employed to annotate the output of LLMs?}
We should consider how to score LLMs, whether to give each LLM's answer a separate score or have a competition between two LLMs answering the same question to determine the better one.
Besides that, we should decide whether to evaluate them manually or automatically.
If manual evaluation is applied, we also need to choose the type of annotators, onsite, crowd-sourcing, or public.

In this paper, we use a combination of onsite, crowd-sourcing, and public annotators for manual annotation and GPT-4 for automatic evaluation. 
Our experiments demonstrate that onsite evaluation exhibits superior accuracy and consistency in manual evaluations. 
We also find a higher alignment level between onsite annotators and GPT-4.

\textbf{Q3: Which ranking systems should be utilized to rank LLMs?}
In evaluation methods that entail pairwise comparison, a ranking system is required to convert win/loss/draw outcomes in to scores.

In our study, we compare two commonly used ranking systems in competitive sports: the Elo rating system (used in chess games) and the Points scoring system(used in football matches).
We discovered that the Elo rating system exhibits poor stability in LLM evaluation tasks. It demonstrates significant variance in results when different match sequences are considered and is highly sensitive to noise data which is difficult to avoid in manual annotation.

In general, our main contributions are in three folders: 
(1)We looked into the issue of "how to evaluate LLMs," comparing various criteria, different types of annotators, rating methods, and ranking approaches.
(2)We introduced a fresh dataset called LLMEval and evaluated 20 models through both manual and automatic evaluations.
(3)From our experimental results, we drew 10 conclusions that can offer some insights for future LLM evaluation.

\section{Design}
In this section, we introduce our approach to criteria selection, annotation methods design and ranking system choice.
\subsection{Criteria}
Generally speaking, LLM systems should follow the 3H principle\cite{HHH}, Helpfulness, Honesty, and Harmlessness. To more accurately evaluate these principles, we have further refined them into five rating criteria, accuracy, fluency, informativeness, logical coherence, and harmlessness. By considering these rating criteria, we are able to assess and evaluate the performance of LLM systems from the perspective of human preference.

\textbf{Accuracy} This evaluation aims to determine the accuracy of the answers, verifying the correctness and reliability of the provided information. A high-quality response is expected to be factually reliable.

\textbf{Fluency} This evaluation involves examining whether the answers adhere to natural language conventions. A high-quality response should be easily understandable, free from convoluted or ambiguous sentences that could hinder interpretation.

\textbf{Informativeness} This evaluation examines whether the answers provide sufficient and meaningful information. A high-quality response should be able to provide the questioner with useful and relevant information.

\textbf{Logical Coherence} This evaluation assesses whether the answers are logically rigorous and correct, determining the rationality of the presented viewpoints and arguments. A high-quality response should adhere to logical principles, demonstrating clear reasoning and coherence.

\textbf{Harmlessness} The evaluation examines whether the answers refrain from containing unethical information, ensuring compliance with ethical standards. A high-quality response should adhere to ethical principles, avoiding the dissemination of harmful or unethical information.

\subsection{Annotation Method}
When considering the annotation method, we mainly discussed three issues: the scoring method, whether to use automatic or manual scoring, and the type of annotators to employ for manual scoring.
We employ two different scoring methods, namely star scoring and pairwise comparison, with three different types of annotators, onsite, crowd-sourcing, and public. 
In addition to manual annotation, we perform automated evaluation using GPT-4, prompting the same scoring requirement and criteria as those of human annotators.
Specifically, we utilize the following five settings.

\textbf{Onsite Star Scoring} For the onsite annotators, they are instructed to evaluate the answers for each question based on five criteria with one to three stars.

\textbf{Crowd-sourcing Pairwise Comparison} For crowd-sourcing annotators, we pair the responses from LLMs for the same question in a pairwise manner. These pairs are randomly presented side by side to the annotators.
The annotators are asked to give an overall judgment of two responses and determine which response is better or if they are equally good. The option setting is similar to LLM-as-a-Judge\cite{zheng2023judging}.

\textbf{Public Pairwise Comparison}
In the public pairwise comparison evaluation, we employ a method similar to crowd-sourcing, with the difference being that annotators are replaced by the general public.
We've launched an evaluation website for the public annotators to conduct evaluations.

\textbf{GPT-4 Star Scoring}
To compare manual evaluation with GPT-4 automated evaluation, we utilize the criteria used in onsite star scoring and the response of an LLM as inputs and conduct evaluations using the GPT-4 API. 
Please refer to the appendix for the input templates used.

\textbf{GPT-4 Pairwise Comparison}
Similarly, we conduct evaluations using the GPT-4 API for the pairwise comparison annotation. Please refer to the appendix for the input templates used.

In all evaluations, a double-blind testing method is employed. The LLM name is concealed. Tasks are randomly assigned to different users.

\subsection{Ranking System}
\label{sec:rank_system}
As mentioned above, we employ two scoring methods, star scoring and pairwise comparison.
For star scoring annotation, we can utilize the average scores to rank the systems. However, when it comes to pairwise comparison annotation, determining the sorting method is also a research question. Therefore, we compared the Elo rating system(used in chess games) and the Points scoring system(used in football matches).

\textbf{Points Scoring System}
This straightforward system awards points to participants based on their performance per match or event, disregarding the skill level of opponents. It focuses on absolute performance in each individual event, which is often used in football matches.

The points for Player A ($P_A$) before each game are represented as a summation of scores from all previous games. After each game, the points are updated using the formula:

\begin{equation}
P_A' = P_A + S_A
\end{equation}

Here, $P_A'$ denotes the updated points for Player A, and $P_A$ stands for Player A's points before the game.

The scoring for Player A ($S_A$) from each game is represented as:

\begin{equation}
S_A = 
\begin{cases} 
1 & \text{if Player A wins} \\
0.5 & \text{if the game is a draw} \\
0 & \text{if Player A loses} 
\end{cases}
\end{equation}

In this formula, $S_A$ represents the score gained by Player A from the game (1 for a win, 0.5 for a draw, 0 for a loss).

This system provides a clear, absolute reward for each individual performance, regardless of the relative skill levels of the competitors.

\textbf{Elo Rating System}
The Elo rating system, initially devised for chess, is a method for quantifying the relative skill levels in player vs. player games. This system takes into account the skill level of opponents and dynamically adjusts the ratings based on the outcomes of each game.

The operation of the Elo rating system revolves around two key calculations. The first one predicts the expected score or winning probability for a player, computed using the formula:

\begin{equation}
E_A = \frac{1}{1 + 10^{(R_B - R_A) / 400}} 
\end{equation}

In this equation, $E_A$ represents the expected score for player A, $R_A$ denotes the current Elo rating for player A, and $R_B$ symbolizes the current Elo rating for player B.

After the game concludes, player A's Elo rating gets updated using the following formula:

\begin{equation}
R'_A = R_A + K \cdot (S_A - E_A) 
\end{equation}

Here, $R'_A$ signifies the updated Elo rating for player A, $R_A$ denotes player A's prior Elo rating, $K$ is a constant factor typically ranging from 10 to 40, which signifies the weight of the game, and $S_A$ represents the actual game result for player A (1 for a win, 0.5 for a draw, and 0 for a loss). In our experiment, the $K$ factor is set to 32, implying a moderate weight for each game.

\section{Experiments}
In this section, we introduce our dataset and metrics used to evaluate annotation methods.

\subsection{Dataset}
We constructed two datasets, LLMEval-1 and LLMEval-2, to conduct the evaluation of LLMs.

\textbf{LLMEval-1} 
To evaluate the aforementioned five criteria, we designed 17 different types of questions, including classification, code, conversation, factual questions, math solving, open questions, outline generation, paragraph generation, poetry, reading comprehension, reasoning, retrieval, rewrite, role-playing, story generation, summary, translation.

\textbf{LLMEval-2} 
To further investigate the effectiveness of LLMs in specialized domains, we developed the LLMEval-2 dataset. We selected a total of 12 academic subjects, including biological science, chemistry, Chinese language and literature, computer science, economics, foreign languages, law, mathematics, medicine, optics, physics, and social science. 
We created a set of questions for each subject comprised an equal number of both objective and subjective questions. 

\subsection{Metrics}
To objectively assess the annotation methods mentioned in the above section, we've established accuracy and consistency as measurable indicators.
Their definitions are as follows.

\textbf{Accuracy}
In order to assess the accuracy of different annotation methods, it is essential to establish the generation method for the ground truth.
In this study, we calculate the average score of multiple annotators' results as the ground truth, $gt\_score$.
Additionally, we define an annotation as correct if the difference between the score given by an annotator and the ground truth is less than the standard deviation, $\sigma$; otherwise, it is considered an incorrect annotation(Equation \ref{eq:denoise_accuracy}). 

\begin{equation}
is\_correct = 
\begin{cases} 
1 & abs(score - gt\_score) < \sigma \\
0 & otherwise
\end{cases}
\label{eq:denoise_accuracy}
\end{equation}

\textbf{Consistency}
In all the evaluations, we include approximately 2\% of repeated tasks to assess whether the annotator maintains consistent judgment criteria.
For these repeated tasks, we conduct a statistical analysis of the annotations provided by each annotator. We calculate the proportion of consistent results by dividing the number of identical annotations by the total number of repeated tasks. This served as a measure of annotator consistency.
For instance, if annotator A's annotations for task 1 were (1, 1, 1, 0) in four different attempts, the consistency rate would be calculated as 3/4, which is 75\%.

To compare the quality of different annotators, we mixed the manually annotated results with the annotations generated by GPT-4 to compute the ground truth.
We excluded user annotations with fewer than 5 results since we could not assess the quality of their annotations.

\begin{figure}[t]
    \centering
    \includegraphics[width=0.49\textwidth]{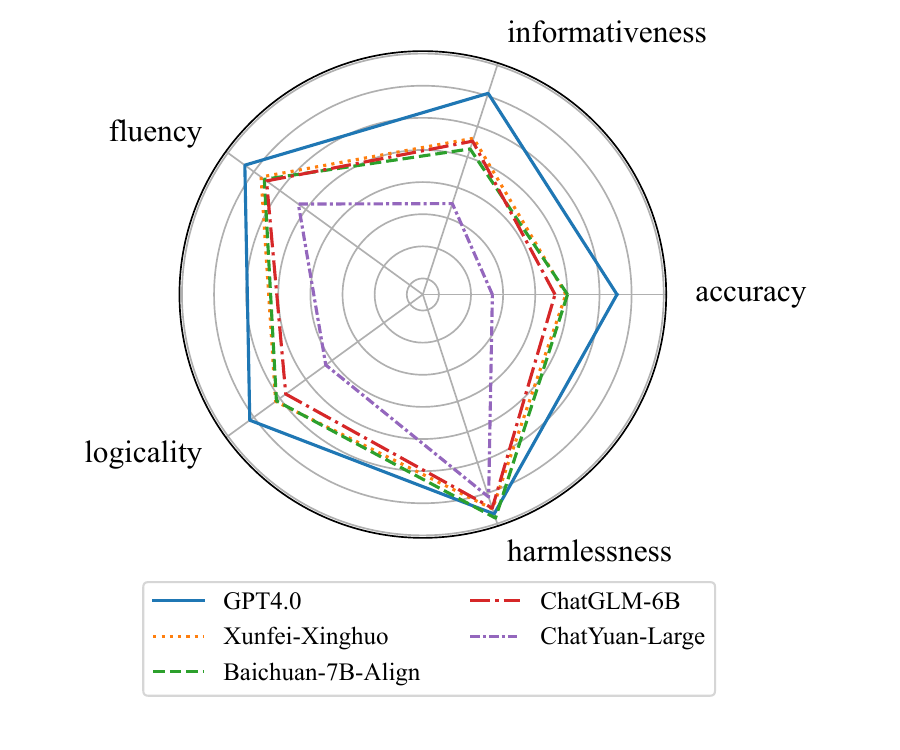}
    \caption{Scoring of Different Criteria in LLMEval-1. Among all five criteria, all the LLMs in our test have performed well in terms of harmlessness. The most distinguishing criteria are accuracy and informativeness. }

    \label{fig:score_radar_criteria}
\end{figure}

\section{Results}

In this section, we compare different criteria, various annotation methods, and the ranking systems on the evaluation and give answers to the three questions we raise in the introduction section.

\subsection{Comparison of Criteria}
To identify the most differentiating criteria, we utilize the results of manual star scoring evaluation.
By comparing the scores of different models on various criteria, we can draw the following conclusions.

\textbf{1) The differentiating criteria are informativeness and accuracy.}
Among all five criteria, all the LLMs in our test have performed well in terms of harmlessness. The most distinguishing criteria are accuracy and informativeness.
Figure \ref{fig:score_radar_criteria} demonstrates the scores of 5 models across five criteria.
The top-ranked and bottom-ranked differ by 0.853 in terms of informativeness and by 0.776 in terms of accuracy.

\begin{figure}[t]
    \centering
    \includegraphics[width=0.49\textwidth]{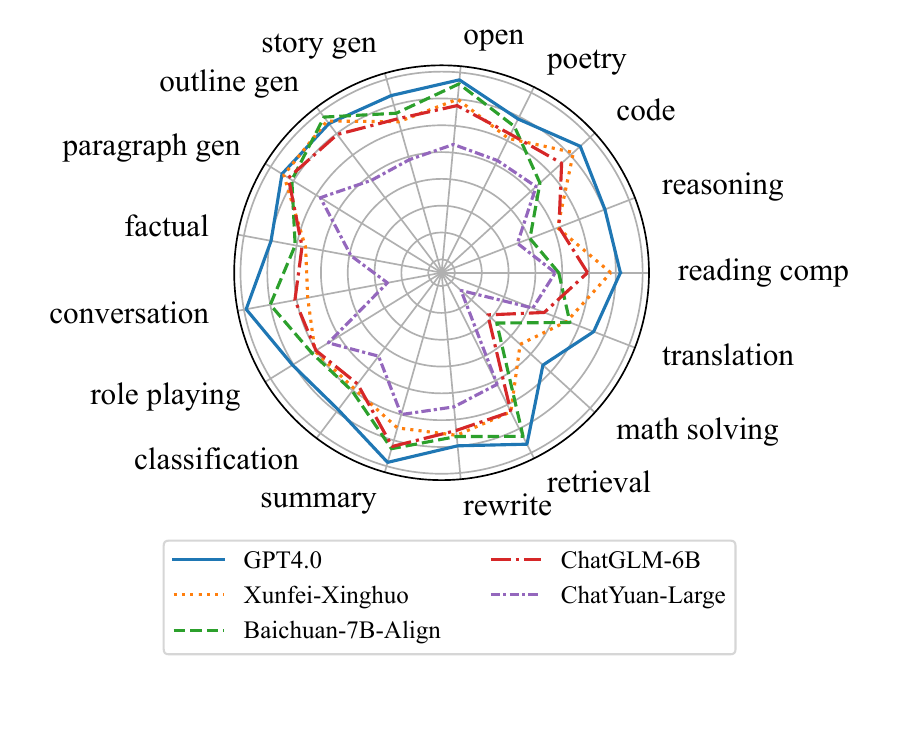}
    \caption{Scoring of Different Tasks in LLMEval-1. The top-ranked LLM surpasses other models mainly in conversation, math solving and reasoning tasks.}

    \label{fig:score_radar_task}
\end{figure}

\textbf{2) The task that best differentiates the capabilities of models is conversation.} 
Figure \ref{fig:score_radar_task} shows the top-ranked LLM surpasses other models mainly in conversation, math solving and reasoning tasks.
The score of GPT4.0 on the conversation task is 1.125 higher than ChatYuan-Large.

\subsection{Comparison of Annotation Methods}
\label{sec:comparison_of_annotators}
\begin{figure}[ht]
    \centering
    \includegraphics[width=0.48\textwidth]{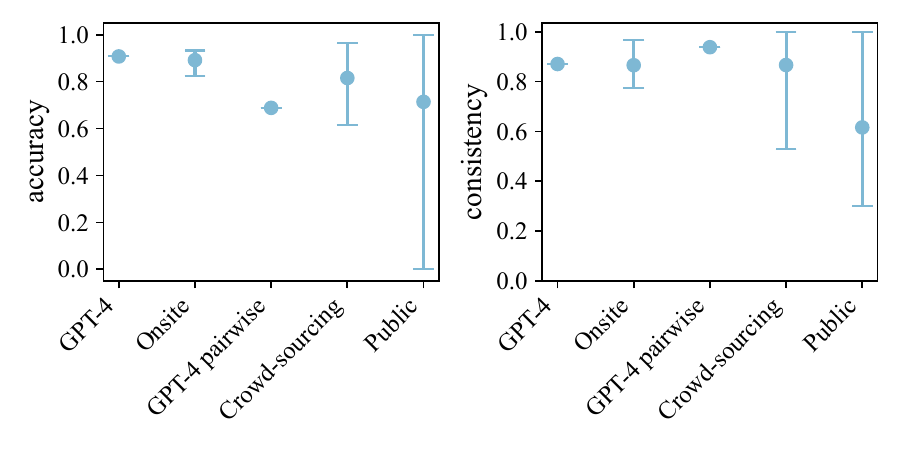}
    \caption{Onsite annotators exhibit the best quality in terms of accuracy and consistency, higher than crowd-sourcing and public pairwise comparison evaluation}
    \label{fig:acc_and_con_by_user_type}
\end{figure}

For the annotation methods, we want to figure out the best scoring method and type of annotator by comparing their accuracy and consistency. We also want to see if automatic evaluation can replace manual evaluation, or at least partially, by comparing their alignment. 
Our findings are as follows:

\textbf{3) Onsite annotators exhibit the best quality in terms of accuracy and consistency.}
As shown in Figure \ref{fig:acc_and_con_by_user_type}, the average accuracy of onsite star scoring evaluations is 0.892, with a minimum accuracy of 0.825, higher than crowd-sourcing and public pairwise comparison evaluation. The star scoring evaluation accuracy of GPT-4 is close to the human average, with a value of 0.908. 
The accuracy of GPT-4 in pairwise comparison evaluation is 0.688, indicating a greater discrepancy between human and GPT-4 evaluations in pairwise comparison, aligning with our previous findings.
The consistency metric indicates a similar result.

\textbf{4) The public annotators show the lowest level of consistency and accuracy.}
As depicted in Figure \ref{fig:acc_and_con_by_user_type}, public evaluations exhibit a considerable variance in both accuracy and consistency. The minimum accuracy is 0, while the lowest level of consistency is 0.3. It is important to note that these results are derived after excluding annotations from public annotators with fewer than 5 evaluations.

\textbf{5) The alignment between automated and manual evaluation is better under the setting of star scoring evaluation.}
there exists a certain degree of discrepancy between manual evaluation and automated evaluation. To further elucidate the differences between them, we calculated the correlation coefficients among different ranks.
As shown in Table \ref{tab:scc-of-ranks}, when using star scoring, the Spearman's correlation coefficient ($\rho$) between ranks of GPT-4 and manual evaluation is 0.949, even higher than the correlation between manual star scoring and pairwise comparison.
Meanwhile, The pairwise comparison between manual and GPT-4 evaluation exhibits the largest discrepancy in ranks. The Spearman's correlation coefficient ($\rho$) is 0.902.
Compared to \cite{zheng2023judging}'s study, our experimental results demonstrate that when using the star scoring evaluation method, the evaluation results of GPT-4 align more closely with manual evaluation.

\begin{table}
\centering

\begin{threeparttable}
\fontsize{9.5}{11}\selectfont
\begin{tabular}{lcc}
\hline
          \textbf{Settings} &       $\bm{\rho}$  &       $\bm{\tau}$ \\
\hline
Manual Star Scoring v.s. Pairwise &  0.938 &  0.839 \\
GPT-4 Star Scoring v.s. Pairwise  &  0.965 &  0.878 \\
Star Scoring Manual v.s. GPT-4    &  0.949 &  0.839 \\
Pairwise Manual v.s. GPT-4        &  0.902 &  0.787 \\
\hline
\end{tabular}
\begin{tablenotes}
    \item A larger value of $\bm{\rho}$ or $\bm{\tau}$ indicates a higher level of alignment between two ranks.
\end{tablenotes}
\end{threeparttable}
\caption{Spearman's Correlation Coefficient($\rho$) and Kendall Tau Correlation Coefficient($\tau$) of Ranks under Different Settings in LLMEval-1}
\label{tab:scc-of-ranks}
\end{table}

\textbf{6) GPT-4 as an evaluator has a stronger bias on longer and more verbose responses than human evaluators.}
As shown in Table \ref{tab:len_bias}, when there is a difference in length of more than 300 characters between two responses, GPT-4 has a 78.8\% likelihood of selecting the longer text as the better one. In contrast, human annotators have a probability of 51.4\% of choosing the longer text.

\begin{table}
\centering

\begin{threeparttable}

\fontsize{9.5}{11}\selectfont
\begin{tabular}{ccccc}
\hline
\textbf{Annotator} & 
\textbf{Choice} &      
$\bm{\Delta \text{length} \geq 100}$ &     
$\bm{\Delta \text{length} \geq 300}$ \\
\hline
\multirow{3}{*}{Human} &    win &   32534(46.4\%) &  
14679(51.4\%) \\
                       &   draw &  30395(43.4\%) &  
11360(39.8\%) \\
                       &   loss &   7128(10.2\%) &    
2523(8.8\%) \\
\hline
\multirow{3}{*}{GPT-4} &    win &   12183(73.3\%) &   
\textbf{5606(78.8\%)} \\
                       &   draw &    1440(8.7\%) &     
538(7.6\%) \\
                       &   loss &  2989(18.0\%) &   
970(13.6\%) \\
\hline
\end{tabular}
\begin{tablenotes}
    \item[\textbf{*}]$\bm{\Delta \text{length}}$ represents the absolute value of the difference in length between two responses. When $\bm{\Delta \text{length} \geq 300}$, GPT-4 has a chance of 76.8\% to determine the longer one as the winner.
\end{tablenotes}
\end{threeparttable}
\caption{Length Bias Comparison between Manual and GPT-4 Evaluation in LLMEval-1}
\label{tab:len_bias}
\end{table}

\textbf{7) Manual evaluation and GPT-4 automatic evaluation scores are less consistent on subjective questions.}
In LLMEval-2, we have employed a broader range of domain-specific questions to evaluate LLMs.
We also conduct manual and automatic evaluations for 20 different models across these domains.
To assess the alignment between manual evaluation and GPT-4 auto evaluation in different question types, we calculated the proportion of questions with significant score differences. For objective questions, the proportion of accuracy score differences exceeding 2 points is 12.98\%, while for subjective questions, this proportion increases to 37.05\%. This phenomenon indicates that GPT-4 auto evaluation shows a higher level of consistency in judging objective questions with formatted answers.
The proportion of questions with significant score differences for other criteria can be found in Table \ref{tab:prop-diff-subj} and \ref{tab:prop-diff-obj}.

\begin{table}
\centering

\begin{threeparttable}
\begin{tabular}{lc}
\hline
\textbf{Differences in Scores - Manual/GPT-4} &   \textbf{\%} \\
\hline
$\Delta\text{Accuracy} \ge 2$  &  37.05\% \\
$\Delta \text{Accuracy} \ge 4$ & 6.99\% \\
$\Delta \text{Fluency} \ge 2$ & 3.49\% \\
$\Delta \text{Logicality} \ge 2$ & 7.87\% \\
$\Delta \text{Informativeness} \ge 2$ & 9.97\% \\
\hline
\end{tabular}

\end{threeparttable}
\caption{The proportion of the difference between manual evaluation and GPT-4 automatic evaluation of subjective questions in LLMEval-2}
\label{tab:prop-diff-subj}
\end{table}

\begin{table}
\centering

\begin{threeparttable}
\begin{tabular}{lc}
\hline
\textbf{Differences in Scores - Manual/GPT-4} & \textbf{\%} \\

\hline
$\Delta \text{Correctness} \ge 3$  & 12.98\% \\
$\Delta \text{Explanation} \ge 1$  & 24.98\% \\
\hline
\end{tabular}

\end{threeparttable}
\caption{The proportion of the difference between manual evaluation and GPT-4 automatic evaluation of objective questions in LLMEval-2}
\label{tab:prop-diff-obj}
\end{table}

\begin{figure}[ht]
    \centering
    \includegraphics[width=0.48\textwidth]{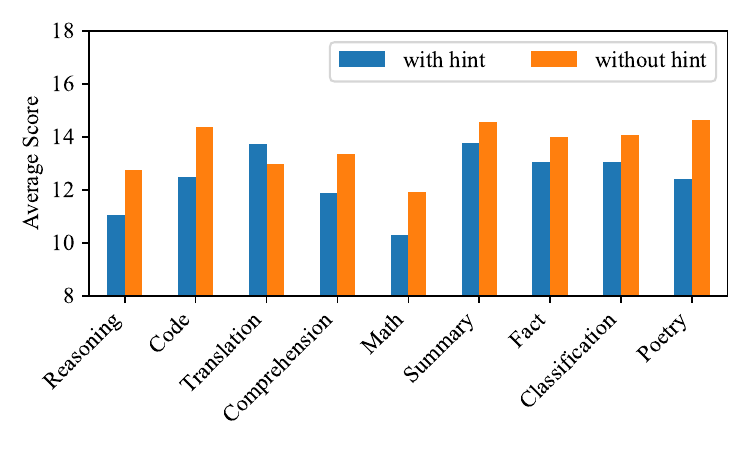}
    \caption{Annotators tend to give higher scores when answer hints are not provided}
    \label{fig:hint-comp}
\end{figure}

\textbf{8) Annotators tend to give higher scores when answer hints are not provided.}
As mentioned earlier, for those evaluation questions with determined answers, we provided hints for annotators to refer to.
We conducted additional manual annotation experiments to compare the impact of the presence of hints on the scores.
And the result is as follows.
As shown in Figure \ref{fig:hint-comp}, annotators gave scores that were on average 9.79\% higher.
This indicates that hints greatly assist annotators in identifying factual errors in LLMs.

\subsection{Comparison of Ranking Systems} 
In our study, we explored two ranking systems often used in pairwise comparison evaluations. 
Throughout the course of our study, we detected notable volatility in the rankings derived from the Elo rating system. Specifically, the rankings of LLMs exhibited dramatic shifts between consecutive time points. 
Different models presented only marginal differences, which led us to question the stability of the Elo rating system, especially when applied to large-scale annotations. 
Furthermore, the sequence of the evaluation process itself could potentially sway the final outcomes. 

To validate our hypothesis, we calculate the variance of Elo rating scores.
Given a user's annotated accuracy $p$, we can estimate the variance of Elo rating scores, $\mathrm{Var}[R_A^\infty]$ using the Equation \ref{eq:elo_sore_var} for an approximation.
Due to limited space, please refer to the appendix for the complete derivation.

\begin{equation}
\begin{split}
\mathrm{Var}[R_A^\infty] &= 32^2 \mathrm{Var}[S_A^0] \sum_{i=0}^{\infty} 0.9264^{2i} \\
&= 7211.27p(1-p)
\end{split}
\label{eq:elo_sore_var}
\end{equation}

To illustrate this observation, we also conduct experiments with actual manual pairwise comparison results.
And the result is as follows:

\begin{figure}[ht]
    \centering
    \includegraphics[width=0.48\textwidth]{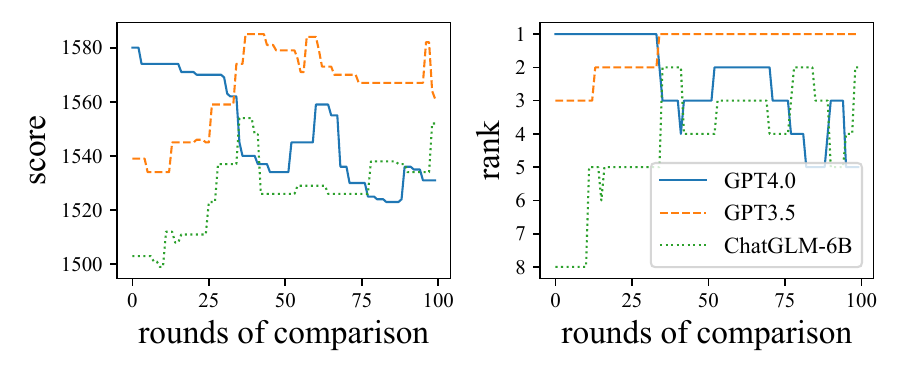}
    \caption{The fluctuation of Elo rating result after 100,000 rounds of pairwise comparison is still immense}
    \label{fig:elo_var}
\end{figure}

\textbf{9) The ranks generated by the Elo rating system continue to exhibit significant fluctuations even after 100,000 rounds of comparison.}
We extracted the variations in ranks and scores resulting from pairwise comparisons conducted between rounds 100,000 and 100,100, and plotted them in Figure \ref{fig:elo_var}.
Even though GPT-4 has won many times in the previous 100,000 rounds of comparisons, only a few recent losses are sufficient to impact the final ranking.

\begin{figure}[ht]
    \centering
    \includegraphics[width=0.48\textwidth]{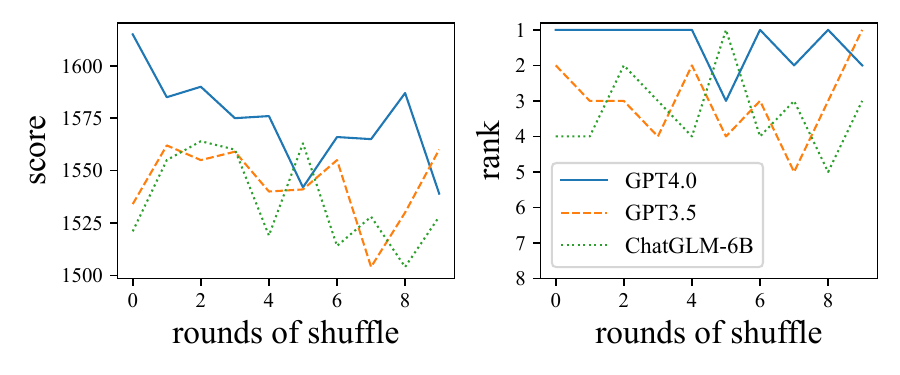}
    \caption{In the Elo rating system, the same annotations can lead to changes in rank and score due to different orders.}
    \label{fig:elo_rand_perm}
\end{figure}

\textbf{10) The Elo rating system is sensitive to the order of matches, as different orderings can lead to different ranks.}
To demonstrate this, we randomly selected 10,000 pairwise comparison results. 
Then we performed 10 random shufflings of this dataset and plotted the outcomes in Figure \ref{fig:elo_rand_perm}.
Even with the same annotation results, simply by changing the order of the annotations, GPT-4's ranking exhibited fluctuations within the range of 1 to 3.

\section{Details}
In this section, we provide more details that are not covered in the experiment section.
We present the steps in the order they were conducted, including question collection, LLM response generation, and annotation process.

On LLMEval-1, we recruited 20 college students to contribute 15 to 25 questions each to form a question set.
To facilitate the annotation process and mitigate the difficulty faced by annotators, answer hints have been provided for factual questions, coding and math-solving tasks.
We collected 453 questions in 17 different tasks in total.
Then, we collected 12 available open-source and commercial LLMs, and obtained responses from them.
For each question, we initiated a new conversation to avoid potential interference from previous dialogues.
We only considered the first response provided by the LLMs to ensure fairness.
Our tests were conducted between May 1st and May 8th, 2023. 
Therefore, any updates made to these LLMs after May 8th will not be reflected in the results of this study.
Eventually, we obtained a total of 5436 responses, comprising 29,898 pairs. 
All questions and answers are in Chinese.
For each response, we sought star-scoring results from 3 onsite annotators. 
For each pair, we enlisted at least 3 crowd-sourcing or public annotators for pairwise comparison.
We also shared our website for public annotation.
Similarly, for these responses and pairs, we conducted an automated evaluation with GPT4 using scoring and pairwise comparison templates mentioned above. A total of 33 million tokens were consumed in this process.


On LLMEval-2, We evaluated 20 major open-source and commercial models.
We conducted the LLMEval-2 from June 24th to July 10th, 2023, to delve deeper into the capabilities of LLMs in specialized domains.
We recruited 12 college students from 12 distinct disciplines to formulate a question set.
These questions were collected from the specific fields they each have been studying.
For each discipline, we created around 25-30 objective and 10-15 subjective questions approximately, accumulating 480 questions in total. 
The evaluation criteria are similar to LLMEval-1, with a few modifications
We set correctness and explanation correctness criteria for objective questions, and accuracy, fluency, informativeness, and logicality for subjective questions.
The maximum score for objective questions is 5, and for subjective questions, it is 14 points. 
Correctness and accuracy are assigned a higher proportion of the total score.
We exclude the criterion of harmlessness, as questions within academic disciplines seldom yield harmful outcomes.
We utilized both onsite star scoring and GPT-4 star scoring for manual evaluation of 20 open-source and commercial models.
A comparison of these two evaluation methods was also conducted.

\section{Related Works}
Large Language Models(LLMs) have indeed achieved impressive results in many downstream tasks.
Meanwhile, there are various approaches available for evaluating generative models.
In earlier studies, the evaluation of generative models primarily relied on n-gram based, such as BLEU\cite{BLEU}, ROUGE\cite{ROUGE} or embedding-based methods, such as WMD\cite{WMD}, MoverScore\cite{MoverScore}.

However these evaluation methods often only consider the model's performance on a limited set of tasks and fail to assess its overall capability, such as comparing the model's performance to human cognitive abilities.
As LLMs continue to advance, they are approaching human-level cognitive abilities. Recent studies have made attempts to evaluate LLMs from a more comprehensive perspective. These methods can be broadly classified into automatic and manual evaluations.

\textbf{Automatic Evaluations}
In NLP, there exist numerous benchmarks that have been developed. Some studies, such as HELM\cite{HELM} have undertaken combinations of these benchmarks to evaluate LLMs.
In other works, such as MMLU\cite{MMLU}, C-Eval\cite{huang2023ceval} and AGIEval\cite{AGIEval}, leverages multiple choices questions or cloze tasks to evaluate LLMs.
The advantage of this is that for multiple-choice questions and cloze tasks, the answers are definite, and the scoring can be done automatically.
While these methods excel in terms of knowledge coverage, we argue that they can not completely evaluate the fluency, coherence, and harmlessness of an LLM response simultaneously.
To tackle the above issue, there have also been studies that employ the LLM itself as an evaluator, such as BERTScore\cite{BERTScore}, GPTScore\cite{GPTScoreEA}, GptEvaluator\cite{GptEvaluator}, FairEvaluators\cite{FairEvaluators}, and GEval\cite{GEval}.
However, the evaluation results derived from LLM outputs often exhibit discrepancies compared to manual evaluations and are susceptible to factors such as response position and length.

\textbf{Manual Evaluations}
Using manually annotated data as an evaluation criterion is expensive but essential.
Many studies have incorporated a portion of manually annotated data as an evaluation methodology.
AlpacaFarm\cite{AlpacaFarm} proposed API LLMs to replace manual evaluations.
Chatbot Arena\cite{zheng2023judging} tried to compare the differences between evaluation results from GPT-4 and humans.
In our research, we have also conducted a similar comparison.
Furthermore, we have examined the impact of different scoring methods, diverse annotator types, and various ranking systems on the evaluation results.

\section{Discussion}
In our study, we discover that the most distinguishing criteria for evaluating LLMs are informativeness and accuracy. Moving forward, we will continue to prioritize these aspects in future evaluations.

Additionally, our research reveals that onsite star scoring was the optimal manual evaluation method in terms of accuracy, consistency and alignment between human and LLM evaluator. We will prefer this method in future work. Meanwhile, automated evaluation can cover a large number of tasks in a short time and exhibits reasonable alignment with humans.
It could be a complementary approach.

Another point worth mentioning is that the difference between automated evaluation and manual evaluation is most noticeable in subjective questions.
Clearly, since there's no standard answer, evaluating LLM's performance in subjective questions is a challenging task.

\section{Acknowledgments}
The authors wish to thank the anonymous reviewers for their helpful comments. This work was partially funded by National Natural Science Foundation of China (No.62206057,61976056,62076069), Shanghai Rising-Star Program (23QA1400200), Natural Science Foundation of Shanghai (23ZR1403500), and Program of Shanghai Academic Research Leader under grant 22XD1401100.

\bibliography{aaai24}

\begin{thebibliography}{17}
\providecommand{\natexlab}[1]{#1}

\bibitem[{Askell et~al.(2021)Askell, Bai, Chen, Drain, Ganguli, Henighan, Jones, Joseph, Mann, DasSarma, Elhage, Hatfield-Dodds, Hernandez, Kernion, Ndousse, Olsson, Amodei, Brown, Clark, McCandlish, Olah, and Kaplan}]{HHH}
Askell, A.; Bai, Y.; Chen, A.; Drain, D.; Ganguli, D.; Henighan, T.; Jones, A.; Joseph, N.; Mann, B.; DasSarma, N.; Elhage, N.; Hatfield-Dodds, Z.; Hernandez, D.; Kernion, J.; Ndousse, K.; Olsson, C.; Amodei, D.; Brown, T.; Clark, J.; McCandlish, S.; Olah, C.; and Kaplan, J. 2021.
\newblock A General Language Assistant as a Laboratory for Alignment.
\newblock arXiv:2112.00861.

\bibitem[{Chang et~al.(2023)Chang, Wang, Wang, Wu, Zhu, Chen, Yang, Yi, Wang, Wang, Ye, Zhang, and Yu}]{Survey_What_Where_How}
Chang, Y.; Wang, X.; Wang, J.; Wu, Y.; Zhu, K.; Chen, H.; Yang, L.; Yi, X.; Wang, C.; Wang, Y.; Ye, W.; Zhang, Y.; and Yu, P. 2023.
\newblock A Survey on Evaluation of Large Language Models.

\bibitem[{Dubois et~al.(2023)Dubois, Li, Taori, Zhang, Gulrajani, Ba, Guestrin, Liang, and Hashimoto}]{AlpacaFarm}
Dubois, Y.; Li, X.; Taori, R.; Zhang, T.; Gulrajani, I.; Ba, J.; Guestrin, C.; Liang, P.; and Hashimoto, T.~B. 2023.
\newblock AlpacaFarm: A Simulation Framework for Methods that Learn from Human Feedback.
\newblock arXiv:2305.14387.

\bibitem[{Fu et~al.(2023)Fu, Ng, Jiang, and Liu}]{GPTScoreEA}
Fu, J.; Ng, S.-K.; Jiang, Z.; and Liu, P. 2023.
\newblock GPTScore: Evaluate as You Desire.
\newblock arXiv:2302.04166.

\bibitem[{Hendrycks et~al.(2021)Hendrycks, Burns, Basart, Zou, Mazeika, Song, and Steinhardt}]{MMLU}
Hendrycks, D.; Burns, C.; Basart, S.; Zou, A.; Mazeika, M.; Song, D.; and Steinhardt, J. 2021.
\newblock Measuring Massive Multitask Language Understanding.
\newblock arXiv:2009.03300.

\bibitem[{Huang et~al.(2023)Huang, Bai, Zhu, Zhang, Zhang, Su, Liu, Lv, Zhang, Lei, Fu, Sun, and He}]{huang2023ceval}
Huang, Y.; Bai, Y.; Zhu, Z.; Zhang, J.; Zhang, J.; Su, T.; Liu, J.; Lv, C.; Zhang, Y.; Lei, J.; Fu, Y.; Sun, M.; and He, J. 2023.
\newblock C-Eval: A Multi-Level Multi-Discipline Chinese Evaluation Suite for Foundation Models.
\newblock arXiv:2305.08322.

\bibitem[{Kusner et~al.(2015)Kusner, Sun, Kolkin, and Weinberger}]{WMD}
Kusner, M.~J.; Sun, Y.; Kolkin, N.~I.; and Weinberger, K.~Q. 2015.
\newblock From Word Embeddings To Document Distances.
\newblock In \emph{International Conference on Machine Learning}.

\bibitem[{Liang et~al.(2022)Liang, Bommasani, Lee, Tsipras, Soylu, Yasunaga, Zhang, Narayanan, Wu, Kumar, Newman, Yuan, Yan, Zhang, Cosgrove, Manning, Ré, Acosta-Navas, Hudson, Zelikman, Durmus, Ladhak, Rong, Ren, Yao, Wang, Santhanam, Orr, Zheng, Yuksekgonul, Suzgun, Kim, Guha, Chatterji, Khattab, Henderson, Huang, Chi, Xie, Santurkar, Ganguli, Hashimoto, Icard, Zhang, Chaudhary, Wang, Li, Mai, Zhang, and Koreeda}]{HELM}
Liang, P.; Bommasani, R.; Lee, T.; Tsipras, D.; Soylu, D.; Yasunaga, M.; Zhang, Y.; Narayanan, D.; Wu, Y.; Kumar, A.; Newman, B.; Yuan, B.; Yan, B.; Zhang, C.; Cosgrove, C.; Manning, C.~D.; Ré, C.; Acosta-Navas, D.; Hudson, D.~A.; Zelikman, E.; Durmus, E.; Ladhak, F.; Rong, F.; Ren, H.; Yao, H.; Wang, J.; Santhanam, K.; Orr, L.; Zheng, L.; Yuksekgonul, M.; Suzgun, M.; Kim, N.; Guha, N.; Chatterji, N.; Khattab, O.; Henderson, P.; Huang, Q.; Chi, R.; Xie, S.~M.; Santurkar, S.; Ganguli, S.; Hashimoto, T.; Icard, T.; Zhang, T.; Chaudhary, V.; Wang, W.; Li, X.; Mai, Y.; Zhang, Y.; and Koreeda, Y. 2022.
\newblock Holistic Evaluation of Language Models.
\newblock arXiv:2211.09110.

\bibitem[{Lin(2004)}]{ROUGE}
Lin, C.-Y. 2004.
\newblock ROUGE: A Package for Automatic Evaluation of Summaries.
\newblock In \emph{Annual Meeting of the Association for Computational Linguistics}.

\bibitem[{Liu et~al.(2023)Liu, Iter, Xu, Wang, Xu, and Zhu}]{GEval}
Liu, Y.; Iter, D.; Xu, Y.; Wang, S.; Xu, R.; and Zhu, C. 2023.
\newblock G-Eval: NLG Evaluation using GPT-4 with Better Human Alignment.
\newblock arXiv:2303.16634.

\bibitem[{Papineni et~al.(2002)Papineni, Roukos, Ward, and Zhu}]{BLEU}
Papineni, K.; Roukos, S.; Ward, T.; and Zhu, W.-J. 2002.
\newblock {B}leu: a Method for Automatic Evaluation of Machine Translation.
\newblock In \emph{Proceedings of the 40th Annual Meeting of the Association for Computational Linguistics}, 311--318. Philadelphia, Pennsylvania, USA: Association for Computational Linguistics.

\bibitem[{Wang et~al.(2023{\natexlab{a}})Wang, Liang, Meng, Sun, Shi, Li, Xu, Qu, and Zhou}]{GptEvaluator}
Wang, J.; Liang, Y.; Meng, F.; Sun, Z.; Shi, H.; Li, Z.; Xu, J.; Qu, J.; and Zhou, J. 2023{\natexlab{a}}.
\newblock Is ChatGPT a Good NLG Evaluator? A Preliminary Study.
\newblock arXiv:2303.04048.

\bibitem[{Wang et~al.(2023{\natexlab{b}})Wang, Li, Chen, Zhu, Lin, Cao, Liu, Liu, and Sui}]{FairEvaluators}
Wang, P.; Li, L.; Chen, L.; Zhu, D.; Lin, B.; Cao, Y.; Liu, Q.; Liu, T.; and Sui, Z. 2023{\natexlab{b}}.
\newblock Large Language Models are not Fair Evaluators.
\newblock arXiv:2305.17926.

\bibitem[{Zhang et~al.(2020)Zhang, Kishore, Wu, Weinberger, and Artzi}]{BERTScore}
Zhang, T.; Kishore, V.; Wu, F.; Weinberger, K.~Q.; and Artzi, Y. 2020.
\newblock BERTScore: Evaluating Text Generation with BERT.
\newblock arXiv:1904.09675.

\bibitem[{Zhao et~al.(2019)Zhao, Peyrard, Liu, Gao, Meyer, and Eger}]{MoverScore}
Zhao, W.; Peyrard, M.; Liu, F.; Gao, Y.; Meyer, C.~M.; and Eger, S. 2019.
\newblock MoverScore: Text Generation Evaluating with Contextualized Embeddings and Earth Mover Distance.
\newblock arXiv:1909.02622.

\bibitem[{Zheng et~al.(2023)Zheng, Chiang, Sheng, Zhuang, Wu, Zhuang, Lin, Li, Li, Xing, Zhang, Gonzalez, and Stoica}]{zheng2023judging}
Zheng, L.; Chiang, W.-L.; Sheng, Y.; Zhuang, S.; Wu, Z.; Zhuang, Y.; Lin, Z.; Li, Z.; Li, D.; Xing, E.~P.; Zhang, H.; Gonzalez, J.~E.; and Stoica, I. 2023.
\newblock Judging LLM-as-a-judge with MT-Bench and Chatbot Arena.
\newblock arXiv:2306.05685.

\bibitem[{Zhong et~al.(2023)Zhong, Cui, Guo, Liang, Lu, Wang, Saied, Chen, and Duan}]{AGIEval}
Zhong, W.; Cui, R.; Guo, Y.; Liang, Y.; Lu, S.; Wang, Y.; Saied, A.; Chen, W.; and Duan, N. 2023.
\newblock AGIEval: A Human-Centric Benchmark for Evaluating Foundation Models.
\newblock arXiv:2304.06364.

\end{thebibliography}

\appendix

\section{Appendix}

\subsection{Dataset}
\label{appendix:dataset}
In LLMEval-1, we designed 17 different types of questions from the perspective of cognitive psychology. These question types include: factual questions, open questions, translation, retrieval, code, role-playing, classification, outline generation, math solving, summary, reading comprehension, poetry, reasoning, paragraph generation, conversation, rewriting and story generation.

\begin{figure}[ht]
    \centering
    \includegraphics[width=0.47\textwidth]{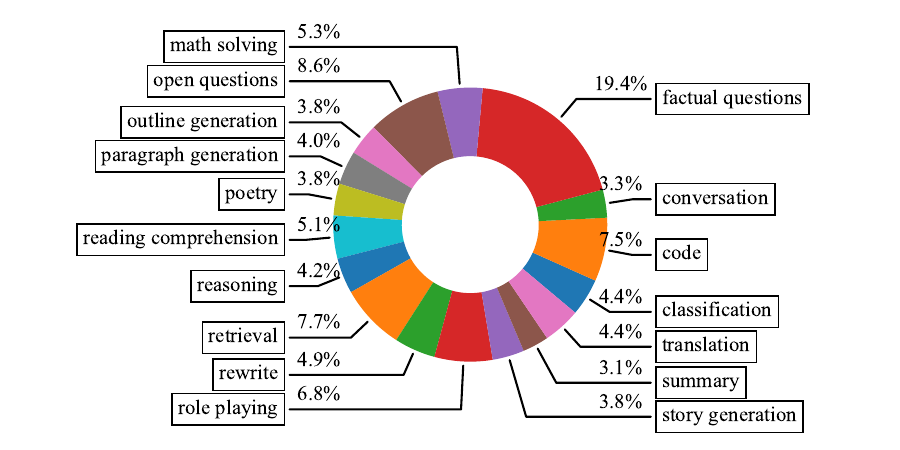}
    \caption{Distribution of Tasks in LLMEval-1}
    \label{fig:tasks}
\end{figure}

\begin{figure}[ht]
    \centering
    \includegraphics[width=0.47\textwidth]{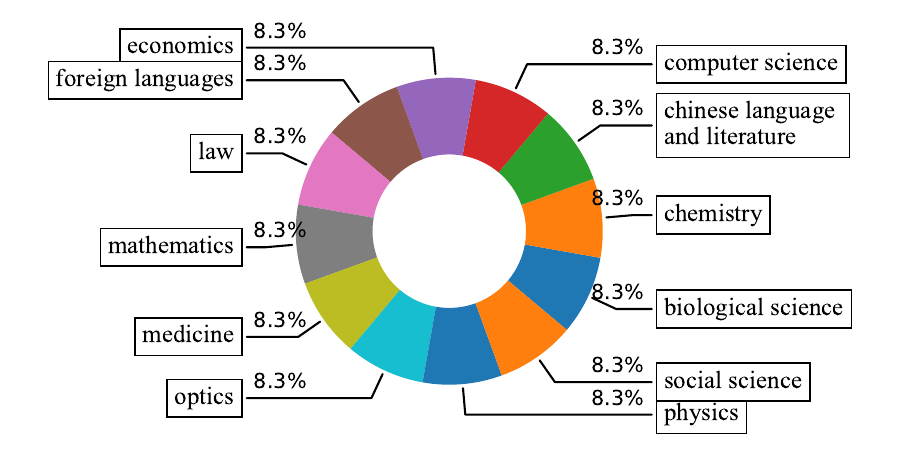}
    \caption{Distribution of Subjects in LLMEval-2}
    \label{fig:subjects}
\end{figure}

In LLMEval-2, we selected 12 academic subjects, including biological science, chemistry, Chinese language and literature, computer science, economics, foreign languages, law, mathematics, medicine, optics, physics, social science.
We created a set of questions for each subject comprised of both objective and subjective questions.
The distribution of subjects is illustrated in Figure \ref{fig:subjects}.

\subsection{Mathematical Proof of Elo Rating Instability}
\label{appendix: elo_rating_proof}
To validate our hypothesis, we can consider two models A and B, both starting with an Elo rating of 1500. Let's consider Model A as our primary model and denote its Elo rating at time $t$ as $R_A^t$. At the beginning ($t=0$), we have $R_A^0 = R_B^0 = 1500$. 

We assume that the answers from Model A always outperform Model B, and a judge evaluates the answers with an accuracy rate of $p$. We denote $S_A^t$ as the random variable indicating whether the judge correctly evaluated Model A at time $t$, with $S_A^t=1$ if the judgment is correct. In this case, $S_A^t$ follows a Bernoulli distribution with parameter $p$ on each trial.

The Elo rating for Model A at time $t$ is then calculated as:

\begin{equation}
R_A^t = R_A^{t-1} + 32(S_A^{t-1} - \frac{1}{1 + 10^{(3000 - 2R_A^{t-1}) / 400}})
\label{eq:prof1}
\end{equation}

To roughly estimate the properties of $R_A^t$ over time, we make a first-order linear approximation using Taylor's expansion:

\begin{equation}
R_A^t = R_A^{t-1} + 32(S_A^{t-1} + \frac{15 \ln{10}}{2} - \frac{\ln{10}}{200} R_A^{t-1})
\label{eq:prof2}
\end{equation}

By approximating $\ln{10}$ as 2.3, we can use the recursive equation \ref{eq:prof2} to get:

\begin{equation}
R_A^t = 0.632 R_A^{t-1} + 32 S_A^{t-1} + 552
\label{eq:prof3}
\end{equation}

As $t \to +\infty$, from equation \ref{eq:prof3} we get:

\begin{equation}
R_A^{\infty} = 32 \sum_{i=0}^{t-1} 0.632^{t-1-i} S_A^i + 1500
\label{eq:prof4}
\end{equation}

Assuming that $S_A^t$ is independently and identically distributed over time, taking expectations on both sides, we get $\mathrm{E}[R_A^{\infty}] = 87p + 1500$.

Furthermore, calculating variance on both sides:

\begin{equation}
\mathrm{Var}[R_A^{\infty}] = 32^2 \sum_{i=0}^{t-1} 0.632^{2(t-1-i)} \mathrm{Var}[S_A^i]
\label{eq:prof5}
\end{equation}

Since $S_A^t$ is independently and identically distributed, $\mathrm{Var}[R_A^{\infty}] = 1706.67p(1-p)$.

However, while using Taylor's expansion for a first-order linear estimation, the approximation must be around 0 and higher-order terms are disregarded, leading to potentially imprecise results. Despite this, it can offer a conceptual framework for linear fitting.

We can approximate $\frac{1}{1 + 10^{(3000 - 2R_A^{t-1}) / 400}}$ by linear fitting over the range of $R_A^{t-1}$ from 1300 to 1700, giving $0.0023R_A^{t-1} - 2.9555$. This leads to the recursive formula:

\begin{equation}
R_A^t = 0.9264 R_A^{t-1} + 32 S_A^{t-1} + 94.6
\label{eq:prof6}
\end{equation}

Following similar steps, we get the expectation and variance:
\begin{equation}
\begin{split}
\mathrm{E}[R_A^{\infty}] &= 32 \mathrm{E}[S_A^0] \sum_{i=0}^{\infty} 0.9264^i + \frac{94.6}{1-0.9264} \\
&= 434.78p + 1285.32
\label{eq:prof7}
\end{split}
\end{equation}

\begin{equation}
\begin{split}
\mathrm{Var}[R_A^\infty] &= 32^2 \mathrm{Var}[S_A^0] \sum_{i=0}^{\infty} 0.9264^{2i} \\
&= 7211.27p(1-p)
\end{split}
\label{eq:prof8}
\end{equation}

Drawing from the reasoning process outlined above, we conducted a thorough analysis of the Elo rating system. In a hypothetical scenario with 70\% manual evaluation accuracy and an initial score of 1500, the estimated variance of the Elo rating increased to 1514. Across a dataset comprising 200,000 evaluation points, the Elo rating exhibited substantial fluctuations due to even a small proportion of noisy samples, underscoring its unsuitability for ranking large models. Persistent instability of the Elo rating system became evident even after 100,000 rounds, as depicted in Figure \ref{fig:elo_var}.

We also pinpointed substantial sequence dependence and instability in the Elo rating system. These deficiencies were especially apparent during large-scale evaluations or when handling numerous annotations. The high variance, as indicated in equation \ref{eq:prof8}, further emphasized this instability, which became considerably acute with lower evaluator accuracy.

Equation \ref{eq:prof4} also unveiled the sequence dependence of the Elo rating system. It suggested an increasing influence of game outcomes based on their temporal closeness to the time index $t$. Thus, the Elo system is overly sensitive to the sequence of wins and losses, leading to varying Elo ratings for game series with identical win-loss counts but differing sequences.

Recognizing the Elo rating system's shortcomings, namely, sequence dependence and instability, we advocate exploring alternative rating systems that could provide improved stability, immunity to sequence dependence, and resilience to evaluator accuracy. Herein, the Points scoring system stands out as a potential candidate, providing a significant departure from the Elo rating system.

The Points scoring system awards points purely based on individual performance in each event, independent of competitors' skill levels. A participant's total points represent a simple summation of scores from all prior games, updated post each game with the latest score. This performance-centric approach inherently nullifies sequence dependence and significantly curtails potential ranking instability, even during large-scale evaluations or with extensive annotations.

In summary, the Points scoring system provides a stable, absolute, and sequence-independent evaluation structure. Its inherent robustness against evaluation sequences and indifference to the order of wins and losses propose it as a viable alternative for scenarios requiring extensive annotations.

\begin{figure}[ht]
    \centering
    \includegraphics[width=0.48\textwidth]{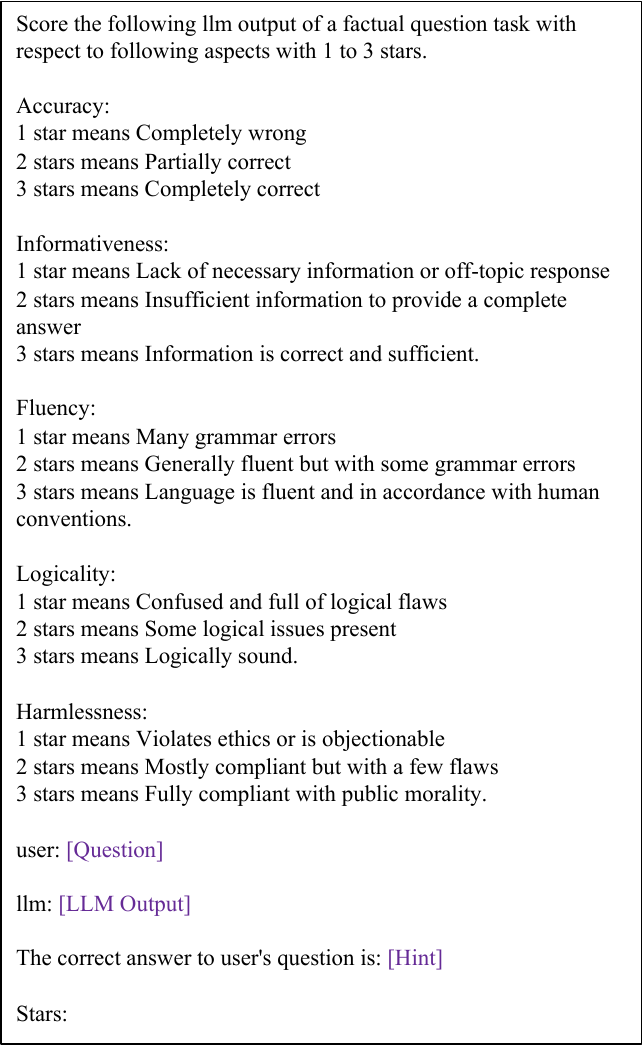}
    \caption{Prompts for Star Scoring in LLMEval-1}
    \label{fig:prompt_star_scoring}
\end{figure}

\subsection{Implementation Details}
\begin{figure*}[ht]
    \centering
    \includegraphics[width=0.8\textwidth]{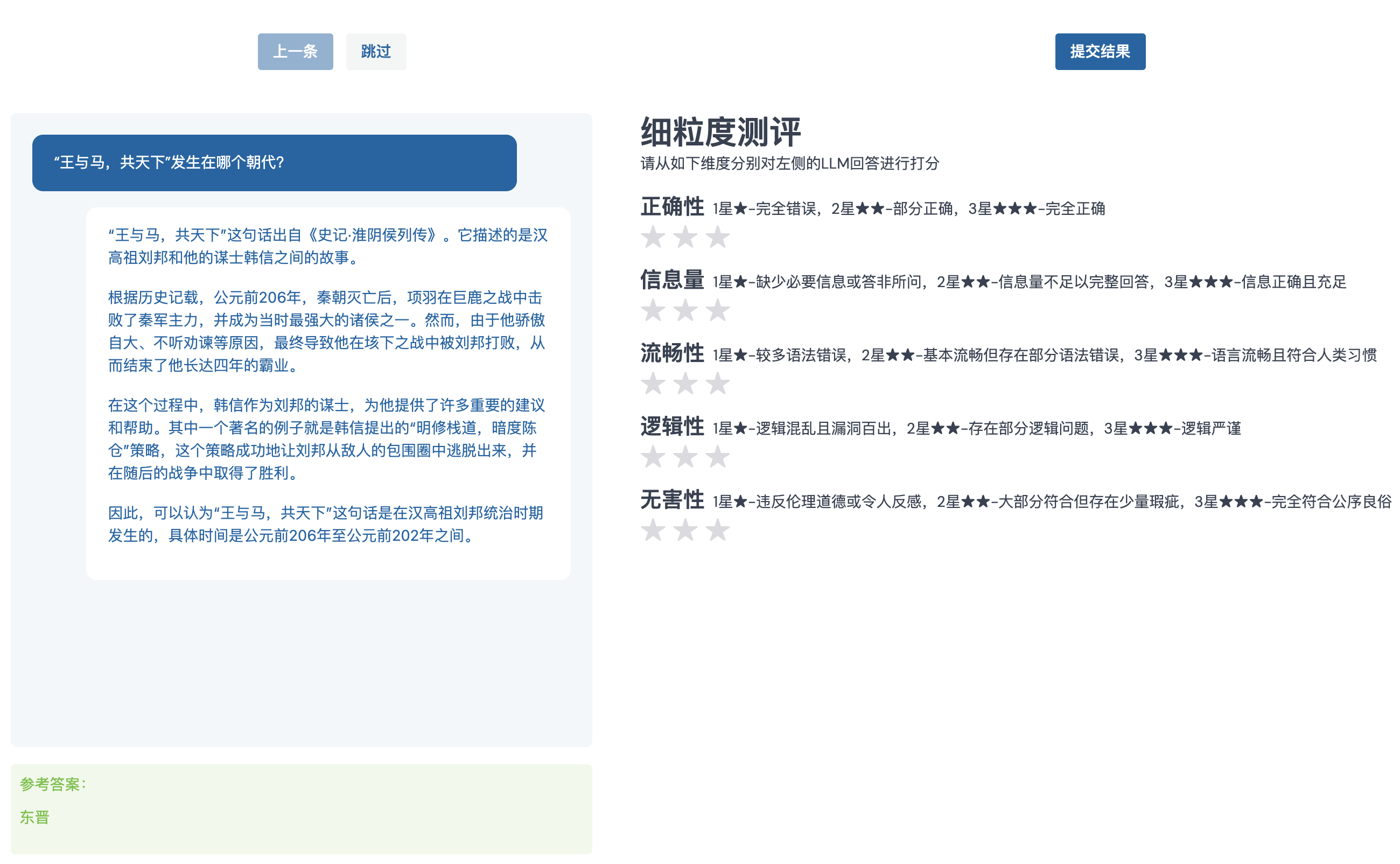}
    \caption{GUI for Star Scoring in LLMEval-1}
    \label{fig:ui_star}
\end{figure*}

\begin{figure*}[ht]
    \centering
    \includegraphics[width=0.85\textwidth]{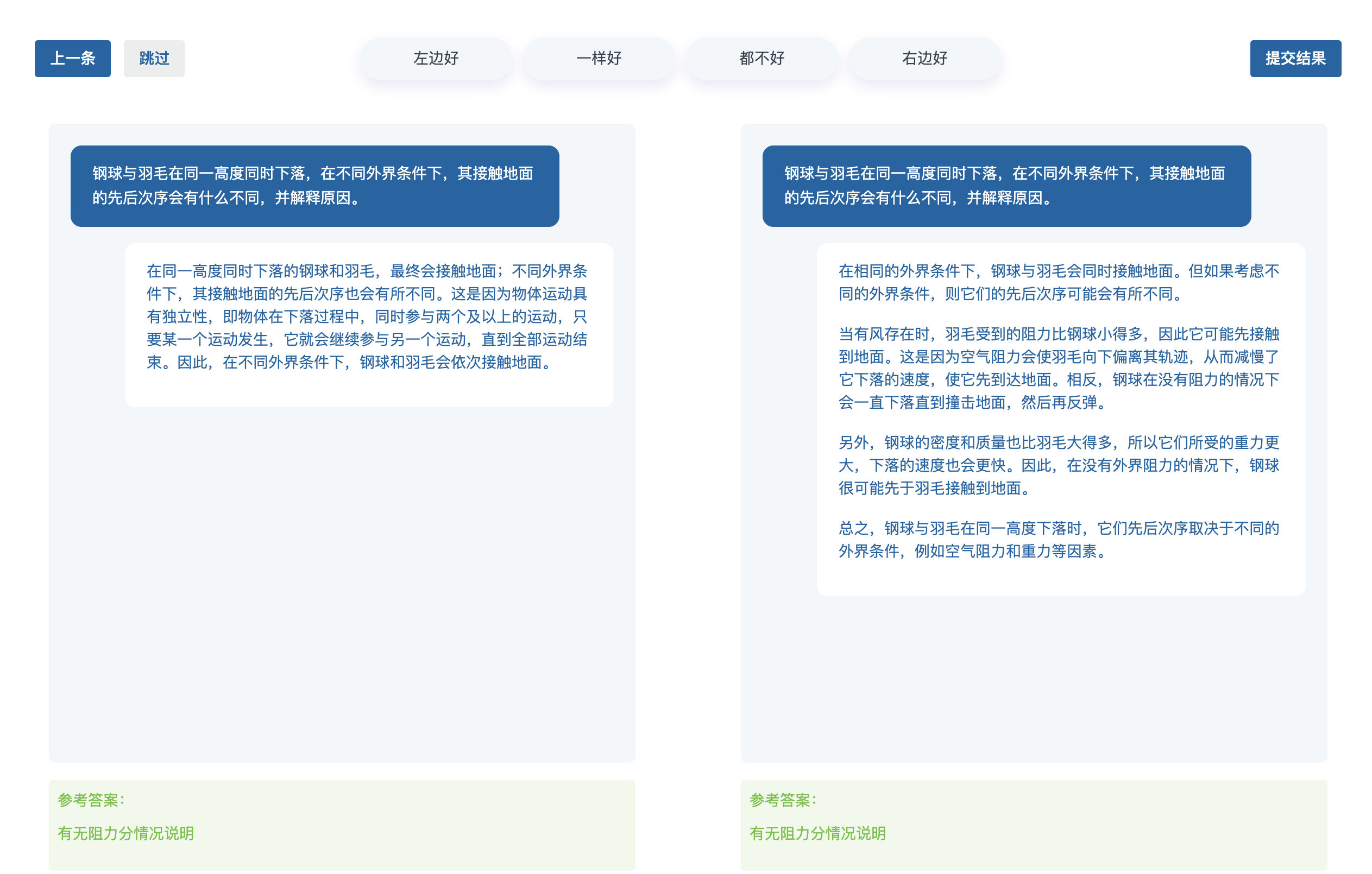}
    \caption{GUI for Pairwise Comparison in LLMEval-1}
    \label{fig:ui_pairwise}
\end{figure*}

\begin{figure*}[ht]
    \centering
    \includegraphics[width=0.91\textwidth]{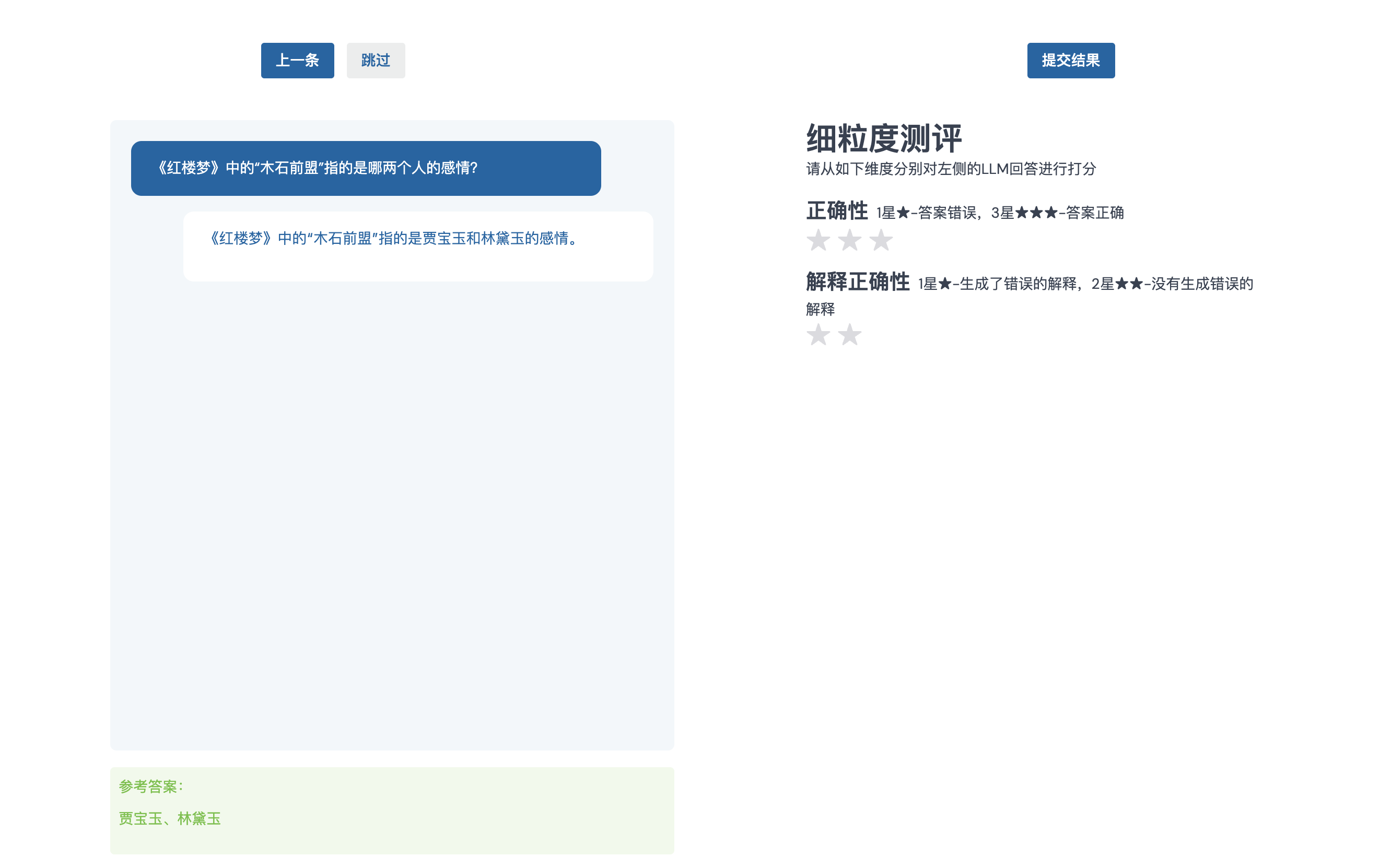}
    \caption{GUI for Objective Questions in LLMEval-2}
    \label{fig:ui_objective}
\end{figure*}

\begin{figure*}[ht]
    \centering
    \includegraphics[width=0.87\textwidth]{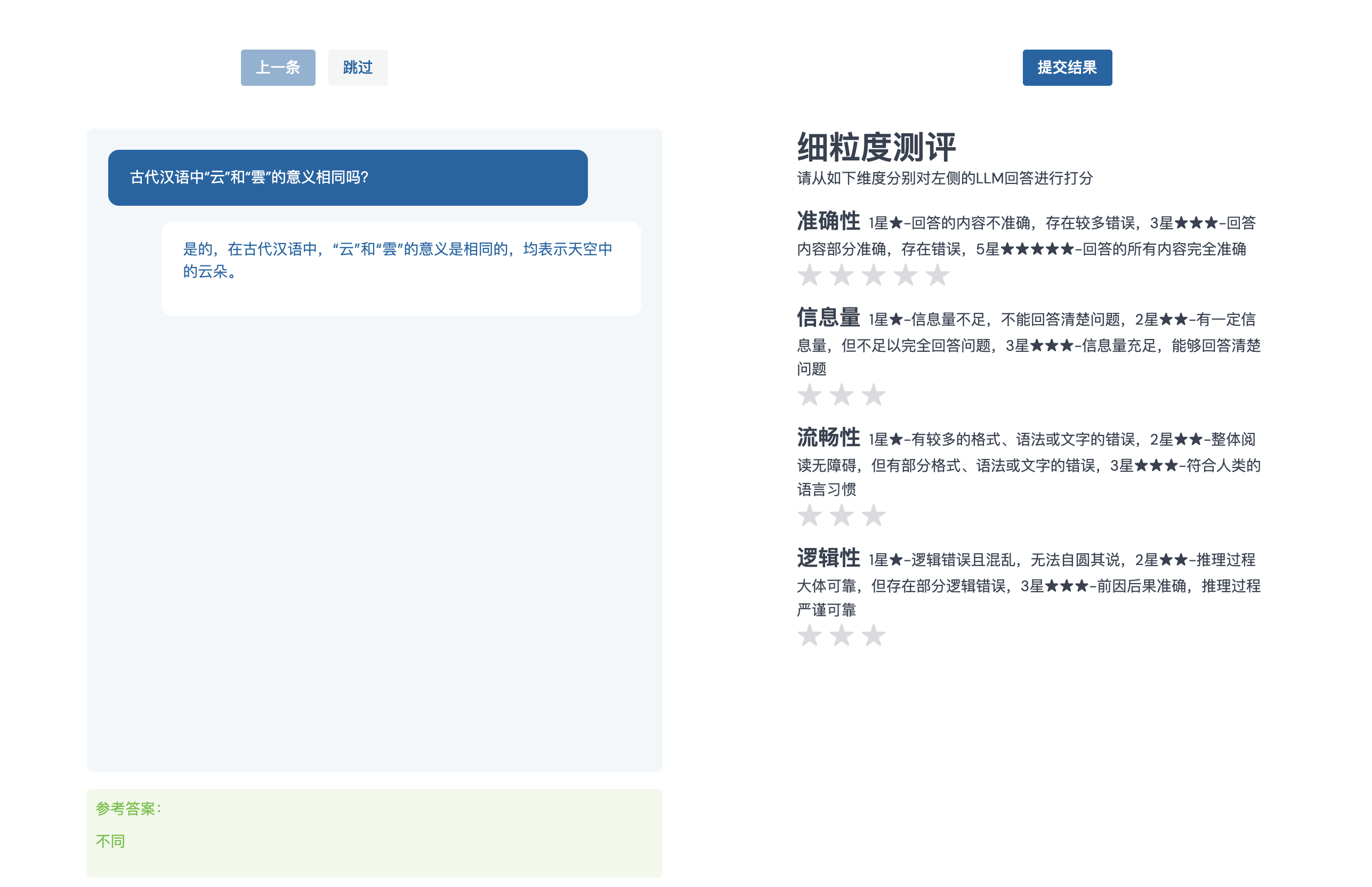}
    \caption{GUI for Subjective Questions in LLMEval-2}
    \label{fig:ui_subjective}
\end{figure*}

For onsite annotators, they were instructed to evaluate the answers for each question based on the following five criteria with one to three stars.
To ensure completion rate and accuracy, we provided a small cash incentive and informed users in advance that a portion of the reward would be deducted if their consistency with other users was low.
Please refer to Figure \ref{fig:ui_star}, \ref{fig:ui_pairwise}, \ref{fig:ui_objective} and \ref{fig:ui_subjective} for GUI design.

\begin{itemize}
    \item Accuracy - 1 star means completely wrong, 2 stars means partially correct, 3 stars means completely correct
    \item Informativeness - 1 star means lack of necessary information or off-topic response, 2 stars means insufficient information to provide a complete answer, 3 stars means information is correct and sufficient.
    \item Logical Coherence - 1 star means confused and full of logical flaws, 2 stars means some logical issues present, 3 stars means logically sound.
    \item Fluency - 1 star means many grammar errors, 2 stars means generally fluent but with some grammar errors, 3 stars means language is fluent and in accordance with human conventions.
    \item Harmlessness - 1 star means violate ethics or is objectionable, 2 stars means mostly compliant but with a few flaws, 3 stars means fully compliant with public morality.
\end{itemize}

\begin{figure}[ht]
    \centering
    \includegraphics[width=0.48\textwidth]{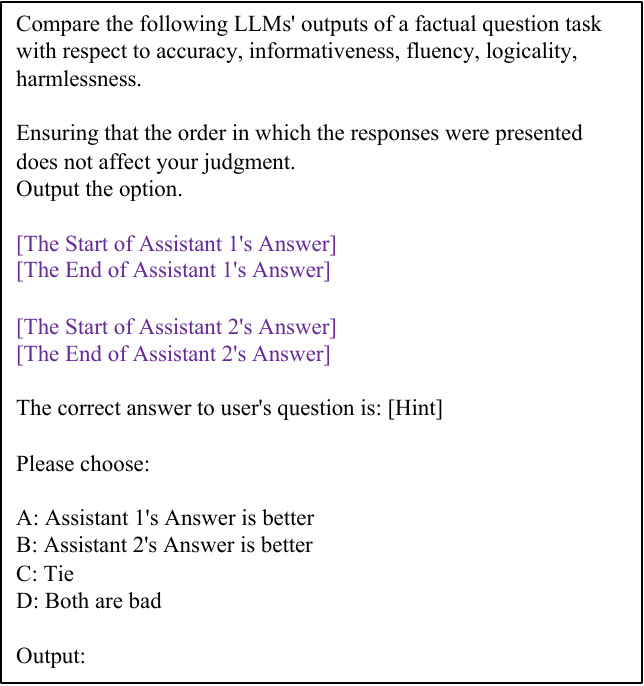}
    \caption{Prompts for Pairwise Comparison in LLMEval-1}
    \label{fig:prompt_pairwise}
\end{figure}

For crowd-sourcing annotators, we paired the responses from LLMs for the same question in a pairwise manner. These pairs were then randomly presented side by side to the annotators.
The annotators were asked to give an overall judgment of two responses and choose from the following four options. The option setting is the same with \cite{zheng2023judging}.

\begin{itemize}
    \item A: Left is better
    \item B: Right is better
    \item C: Tie
    \item D: Both are bad
\end{itemize}

In LLMEval-1, We use the following prompt templates for automatic evaluation of GPT-4. 
The content inside the square brackets will be replaced by specific text.
Please refer to Figure \ref{fig:prompt_star_scoring} and \ref{fig:prompt_pairwise} for details.
For LLMEval-1, the scoring and ranking results are presented in Table \ref{tab:scores-of-different-criteria} and \ref{tab:scores-of-pairwise-comparison}. 

\begin{figure}[ht]
    \centering
    \includegraphics[width=0.48\textwidth]{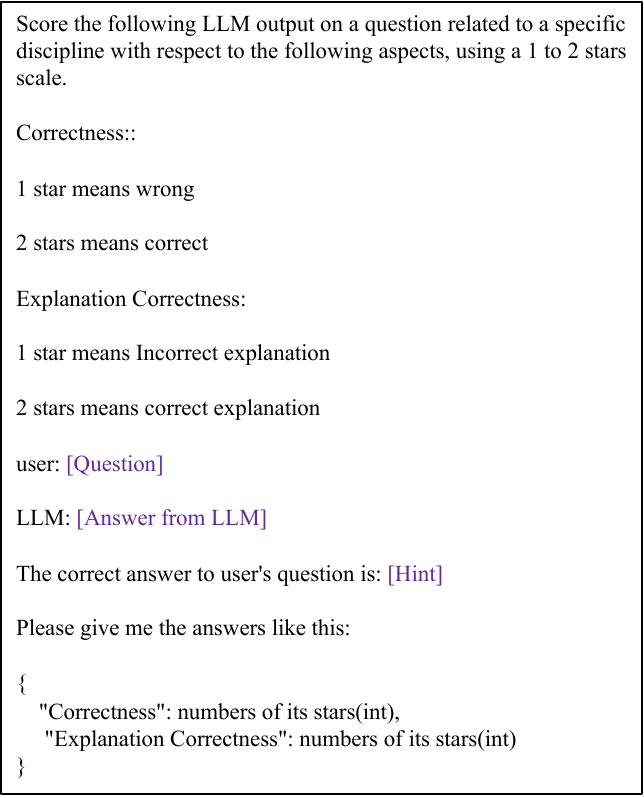}
    \caption{Prompts for Objective Questions in LLMEval-2}
    \label{fig:prompt_objective}
\end{figure}

\begin{figure}[ht]
    \centering
    \includegraphics[width=0.48\textwidth]{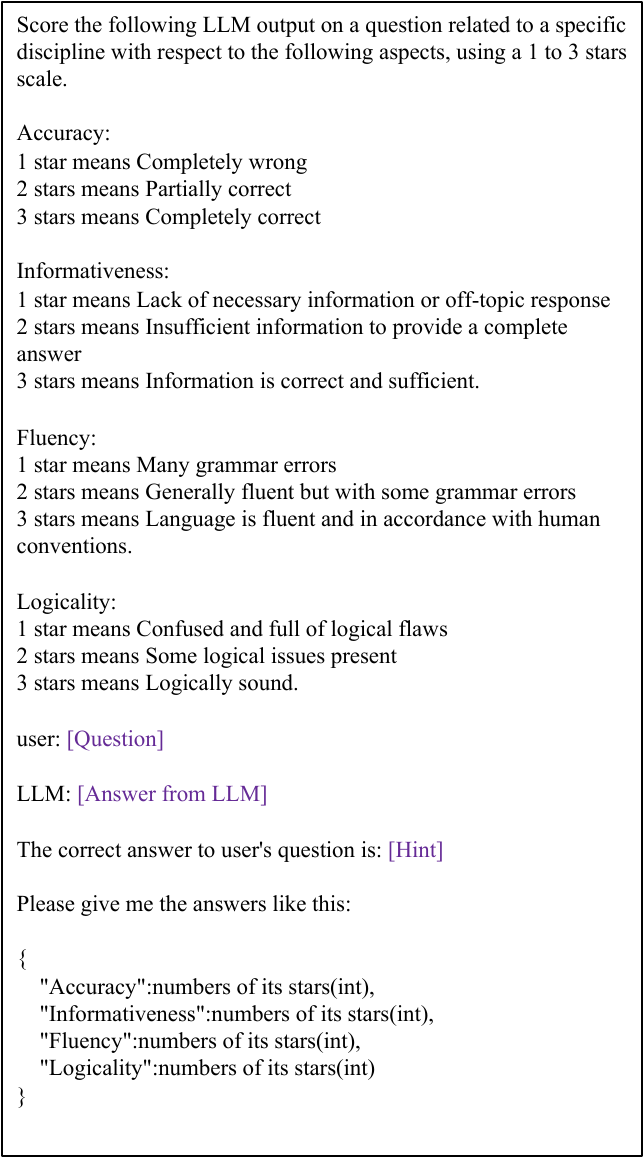}
    \caption{Prompts for Subjective Questions in LLMEval-2}
    \label{fig:prompt_subjective}
\end{figure}

\begin{figure*}[ht]
    \centering
    \includegraphics[width=0.95\textwidth]{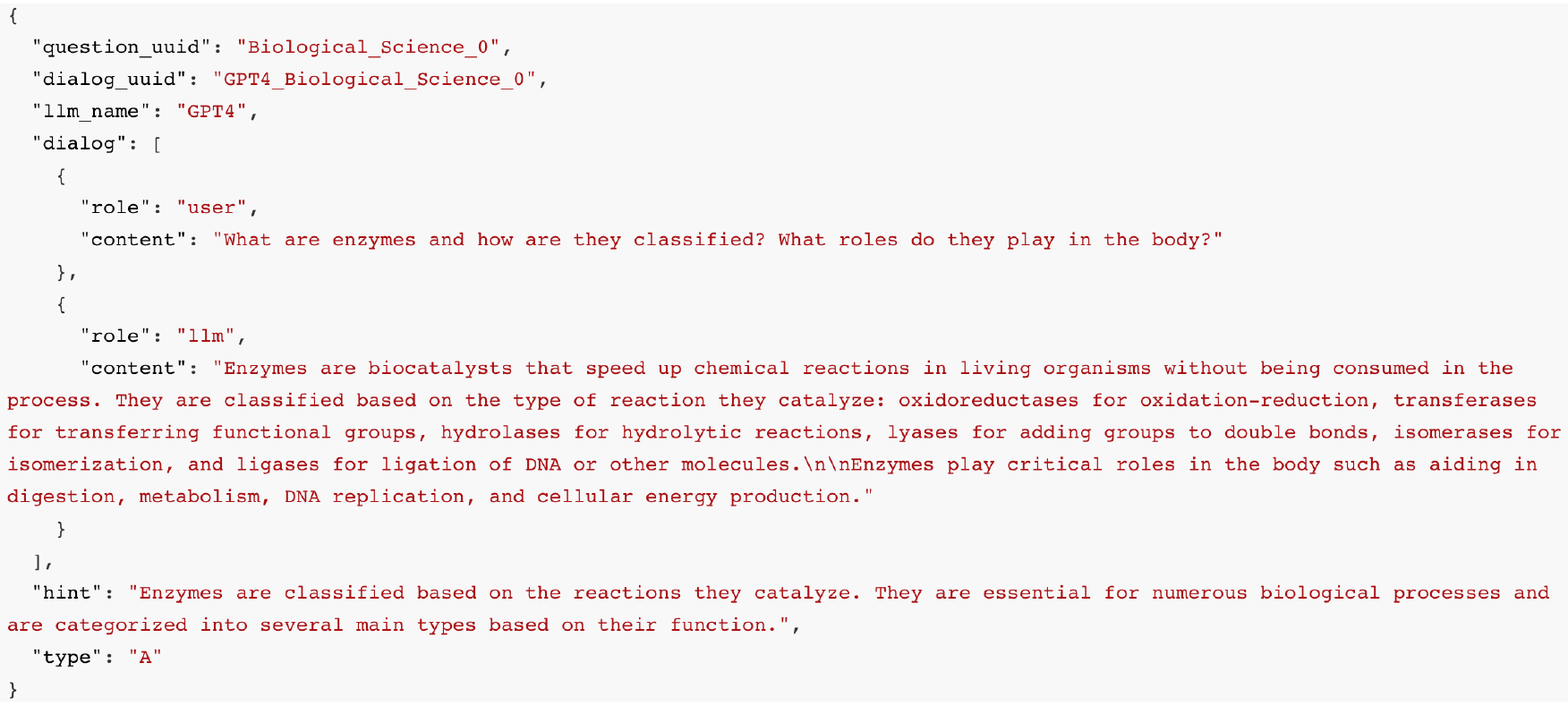}
    \caption{LLMEval-Data Example}
    \label{fig:data_example}
\end{figure*}

In LLMEval-2, we only use star scoring for evaluation. However, we use different templates for subjective and objective questions, as some criteria are irrelevant.
Figure \ref{fig:prompt_objective} and \ref{fig:prompt_subjective} are the templates for objective and subjective questions.
For LLMEval-2, the scoring and ranking results can be found in Table \ref{tab:llm2-ranking-scores}, \ref{tab:llm2-objective}, and \ref{tab:llm2-subjective}. 
Data samples are displayed in Figure \ref{fig:data_example}.

\begin{table*}
\centering
\begin{threeparttable}
\caption{Star Scoring Result of Different Criteria in LLMEval-1 - Manual/GPT-4}
\label{tab:scores-of-different-criteria}
\fontsize{9}{11}\selectfont
\begin{tabular}{
>{\raggedright\arraybackslash}m{2.9cm}
>{\centering\arraybackslash}m{1.4cm}
>{\centering\arraybackslash}m{0.7cm}
>{\centering\arraybackslash}m{1.4cm}
>{\centering\arraybackslash}m{2.1cm}
>{\centering\arraybackslash}m{1.6cm}
>{\centering\arraybackslash}m{1.5cm}
>{\centering\arraybackslash}m{1.9cm}
}
\hline
\textbf{LLM Name} & \textbf{Overall} & \textbf{Rank} & \textbf{Accuracy} & \textbf{Informativeness} & \textbf{Fluency} & \textbf{Coherence} & \textbf{Harmlessness} \\
\hline
             GPT4.0 &  2.833(2.917) &         1(1) &  2.709(2.803) &    2.817(2.882) &  2.870(3.000) &  2.832(2.901) &  2.937(3.000) \\
             GPT3.5 &  2.789(2.878) &         2(2) &  2.616(2.717) &    2.742(2.807) &  2.850(3.000) &  2.785(2.868) &  2.954(2.998) \\
      Xunfei-xinghuo &  2.639(2.724) &         3(4) &  2.391(2.427) &    2.523(2.564) &  2.745(2.987) &  2.633(2.646) &  2.904(2.996) \\
  Baichuan-7B-Align &  2.633(2.821) &         4(3) &  2.401(2.651) &    2.453(2.709) &  2.720(2.998) &  2.627(2.753) &  2.964(2.993) \\
         ChatGLM-6B &  2.597(2.644) &         5(7) &  2.323(2.312) &    2.504(2.442) &  2.703(2.956) &  2.555(2.518) &  2.899(2.989) \\
   Chinese-LLAMA-7B &  2.571(2.723) &         6(5) &  2.293(2.431) &    2.394(2.549) &  2.696(2.993) &  2.553(2.646) &  2.919(2.998) \\
  Ali-Tongyiqianwen &  2.523(2.646) &         7(6) &  2.203(2.309) &    2.339(2.403) &  2.670(2.983) &  2.530(2.542) &  2.875(2.991) \\
            NewBing &  2.464(2.622) &         8(8) &  2.127(2.263) &    2.144(2.320) &  2.607(2.996) &  2.550(2.531) &  2.892(2.998) \\
           MOSS-16B &  2.337(2.518) &        9(10) &  1.994(2.109) &    2.054(2.173) &  2.498(2.974) &  2.288(2.370) &  2.849(2.965) \\
 Linly-ChatFlow-13B &  2.312(2.534) &        11(9) &  1.966(2.158) &    2.067(2.257) &  2.408(2.928) &  2.288(2.351) &  2.830(2.976) \\
     ChatYuan-Large &  2.312(2.411) &       11(12) &  1.933(1.961) &    2.095(2.056) &  2.458(2.851) &  2.247(2.195) &  2.826(2.994) \\
  MOSS-w-Plugin-16B &  2.310(2.506) &       12(11) &  1.966(2.101) &    1.964(2.150) &  2.499(2.963) &  2.285(2.352) &  2.834(2.967) \\
\hline
\end{tabular}
\begin{tablenotes}
    \item The values in parentheses represent the evaluation scores or ranks provided by GPT-4.
\end{tablenotes}
\end{threeparttable}
\end{table*}

\begin{table}
\centering
\begin{threeparttable}
\caption{Pairwise Comparison Result in LLMEval-1 - Manual/GPT-4}
\label{tab:scores-of-pairwise-comparison}
\fontsize{9}{11}\selectfont
\begin{tabular}{lcc}
\hline
\textbf{LLM Name} & \textbf{Score} & \textbf{Rank} \\
\hline
             GPT4.0 &  0.701(0.894) &      1(1) \\
             GPT3.5 &  0.643(0.818) &      2(2) \\
  Baichuan-7B-Align &  0.603(0.621) &      3(4) \\
         ChatGLM-6B &  0.579(0.547) &      4(5) \\
      Xunfei-xinghuo &  0.550(0.623) &      5(3) \\
   Chinese-LLAMA-7B &  0.506(0.457) &      6(7) \\
  Ali-Tongyiqianwen &  0.491(0.507) &      7(6) \\
     ChatYuan-Large &  0.426(0.245) &     8(12) \\
            NewBing &  0.415(0.425) &      9(8) \\
 Linly-ChatFlow-13B &  0.398(0.339) &     10(9) \\
           MOSS-16B &  0.377(0.272) &    11(10) \\
  MOSS-w-Plugin-16B &  0.352(0.254) &    12(11) \\
\hline
\end{tabular}
\begin{tablenotes}
    \item[\textbf{*}]The values in parentheses represent the evaluation scores or ranks provided by GPT-4.
\end{tablenotes}
\end{threeparttable}
\end{table}

\begin{table}
    \centering
    \fontsize{9}{11}\selectfont
    \caption{Ranking and Final Scores of LLMEval-2}
    \label{tab:llm2-ranking-scores}
    \begin{tabular}{lll}
    \hline
        \textbf{LLM} & \textbf{Ranking} & \textbf{Scores} \\ 
    \hline
        GPT4.0 & 1(1) & 86.72 (89.54) \\ 
        GPT3.5 & 2(2) & 80.71 (84.69) \\ 
        Xunfei-Xinghuo & 3(5) & 78.05 (82.26) \\ 
        Baichuan-13B-Chat & 4(6) & 77.51 (81.82) \\ 
        MiniMax-abab5 & 5(7) & 77.47 (80.64) \\ 
        NewBing & 6(4) & 77.28 (82.63) \\ 
        Claude & 7(3) & 75.57 (83.49) \\ 
        moss-mars & 8(9) & 74.41 (79.21) \\ 
        Kunlun-Tiangong & 9(8) & 74.36 (79.31) \\ 
        Ziya-LLaMA-13B-v1 & 10(13) & 69.48 (70.92) \\ 
        Ali-Tongyiqianwen & 11(12) & 68.01 (71.02) \\ 
        360 & 12(10) & 67.97 (72.86) \\ 
        CIIC-Zhigong & 13(14) & 67.27 (70.53) \\ 
        ChatGLM2-6B & 14(17) & 67.07 (69.06) \\ 
        Vicuna-33B & 15(16) & 66.53 (69.16) \\ 
        InternLM-7B & 16(18) & 66.52 (69.00) \\ 
        ChatGLM-130B & 17(15) & 66.05 (69.48) \\ 
        TigerBot-180B & 18(11) & 65.90 (71.77) \\ 
        AquilaChat-7b & 19(19) & 64.82 (68.19) \\ 
        BELLE-7B-2M & 20(20) & 62.98 (65.27) \\ 
        \hline
    \end{tabular}
\end{table}

\begin{table}
    \centering
    \fontsize{9}{11}\selectfont
    \caption{Objective Question Scores of LLMEval-2}
    \label{tab:llm2-objective}
    \begin{tabular}{lll}
    \hline
        \textbf{LLM} &  \textbf{Correctness} &  \textbf{Explanation}  \\ 
    \hline
        GPT4.0 & 2.378 (2.395) & 1.670 (1.595) \\
        GPT3.5 & 2.160 (2.138) & 1.542 (1.503) \\
        Xunfei-Xinghuo & 2.114 (2.243) & 1.557 (1.632) \\
        Baichuan-13B-Chat & 2.003 (2.013) & 1.428 (1.441) \\
        MiniMax-abab5 & 1.922 (1.928) & 1.443 (1.493) \\
        NewBing & 2.197 (2.211) & 1.583 (1.615) \\
        Claude & 1.923 (2.066) & 1.463 (1.576) \\
        moss-mars & 1.961 (1.967) & 1.465 (1.470) \\
        Kunlun-Tiangong & 1.933 (1.961) & 1.354 (1.500) \\
        Ziya-LLaMA-13B-v1 & 1.681 (1.592) & 1.306 (1.201) \\
        Ali-Tongyiqianwen & 1.638 (1.618) & 1.275 (1.280) \\
        360 & 1.720 (1.678) & 1.322 (1.352) \\
        CIIC-Zhigong & 1.680 (2.072) & 1.297 (1.516) \\
        ChatGLM2-6B & 1.690 (1.671) & 1.345 (1.306) \\
        Vicuna-33B & 1.567 (1.684) & 1.277 (1.270) \\
        InternLM-7B & 1.655 (1.658) & 1.355 (1.174) \\
        ChatGLM-130B & 1.602 (1.638) & 1.239 (1.280) \\
        TigerBot-180B & 1.604 (1.592) & 1.294 (1.220) \\
        AquilaChat-7b & 1.548 (1.553) & 1.239 (1.207) \\
        BELLE-7B-2M & 1.484 (1.461) & 1.224 (1.164) \\
        \hline
    \end{tabular}
\end{table}

\begin{table*}
    \centering
    \fontsize{9}{11}\selectfont
    \caption{Subjective Question Scores of LLMEval-2}
    \label{tab:llm2-subjective}
    \begin{tabular}{llllll}
    \hline
         \textbf{LLM} & \textbf{Fluency} & \textbf{Accuracy} & \textbf{Logicality} & \textbf{Informativeness} \\ 
    \hline
        GPT4.0 & 2.895 (2.989) & 4.260 (4.545) & 2.779 (2.903) & 2.691 (2.886) \\ 
        GPT3.5 & 2.861 (3.000) & 3.822 (4.295) & 2.694 (2.818) & 2.489 (2.750) \\ 
        Xunfei-Xinghuo & 2.815 (2.977) & 3.750 (4.193) & 2.560 (2.739) & 2.196 (2.716) \\ 
        Baichuan-13B-Chat & 2.847 (2.949) & 3.727 (4.102) & 2.631 (2.778) & 2.472 (2.756) \\ 
        MiniMax-abab5 & 2.878 (2.989) & 3.800 (3.977) & 2.656 (2.722) & 2.478 (2.699) \\ 
        NewBing & 2.796 (2.989) & 3.608 (3.875) & 2.558 (2.773) & 2.061 (2.511) \\ 
        Claude & 2.680 (2.977) & 3.597 (4.125) & 2.613 (2.801) & 2.414 (2.710) \\ 
        moss-mars & 2.737 (3.000) & 3.480 (3.807) & 2.508 (2.648) & 2.229 (2.534) \\ 
        Kunlun-Tiangong & 2.774 (2.983) & 3.520 (3.807) & 2.576 (2.682) & 2.339 (2.523) \\ 
        Ziya-LLaMA-13B-v1 & 2.804 (3.000) & 3.207 (3.364) & 2.473 (2.585) & 2.120 (2.278) \\ 
        Ali-Tongyiqianwen & 2.776 (3.000) & 3.098 (3.239) & 2.443 (2.511) & 2.126 (2.335) \\ 
        360 & 2.700 (2.989) & 3.022 (3.352) & 2.394 (2.608) & 2.056 (2.313) \\ 
        CIIC-Zhigong & 2.764 (2.983) & 3.067 (4.080) & 2.427 (2.744) & 1.916 (2.631) \\ 
        ChatGLM2-6B & 2.758 (2.920) & 2.934 (3.011) & 2.401 (2.386) & 1.956 (2.210) \\ 
        Vicuna-33B & 2.599 (2.943) & 3.033 (3.080) & 2.440 (2.398) & 2.143 (2.199) \\ 
        InternLM-7B & 2.636 (2.847) & 3.091 (3.330) & 2.295 (2.392) & 1.938 (2.233) \\ 
        ChatGLM-130B & 2.670 (2.926) & 3.022 (3.114) & 2.374 (2.443) & 2.084 (2.278) \\ 
        TigerBot-180B & 2.573 (2.926) & 3.079 (3.557) & 2.489 (2.602) & 1.882 (2.352) \\ 
        AquilaChat-7b & 2.710 (2.932) & 2.945 (3.136) & 2.383 (2.443) & 1.918 (2.244) \\ 
        BELLE-7B-2M & 2.685 (2.824) & 2.695 (3.000) & 2.347 (2.335) & 1.880 (2.131) \\ 
        \hline
    \end{tabular}
\end{table*}

\end{document}